\def\year{2022}\relax
\documentclass[letterpaper]{article}
\usepackage{aaai22} 
\usepackage{times} 
\usepackage{helvet} 
\usepackage{courier} 
\usepackage[hyphens]{url} 
\usepackage{subcaption}
\usepackage{bm}
\usepackage{graphicx}
\usepackage{natbib} 
\usepackage{caption}
\DeclareCaptionStyle{ruled}{labelfont=normalfont,labelsep=colon,strut=off}
\usepackage{algorithm}
\usepackage{algorithmic}

\usepackage{newfloat}
\usepackage{listings}
\usepackage{amsmath}
\floatstyle{ruled}
\newfloat{listing}{tb}{lst}{}
\floatname{listing}{Listing}

\title{A Question on the Explainability of Large Language Models \\ and the Word-Level Univariate First-Order Plausibility Assumption}
\author {
    Jeremie Bogaert\textsuperscript{\rm 1},
    François-Xavier Standaert\textsuperscript{\rm 1}
}
\affiliations {
    \textsuperscript{\rm 1} Crypto Group, ICTEAM Institute, UCLouvain, Belgium\\
    jeremie.bogaert@uclouvain.be, fstandae@uclouvain.be
}

\begin{document}
\maketitle

\begin{abstract} 
The explanations of large language models have recently been shown to be sensitive to the randomness used for their training, creating a need to characterize this sensitivity. In this paper, we propose a characterization that questions the possibility to provide simple and informative explanations for such models. To this end, we give statistical definitions for the explanations' signal, noise and signal-to-noise ratio. We highlight that, in a typical case study where word-level univariate explanations are analyzed with first-order statistical tools, the explanations of simple feature-based models carry more signal and less noise than those of transformer ones. We then discuss the possibility 
to improve these results with alternative definitions of signal and noise  that would capture more complex explanations and analysis methods, while  also questioning the tradeoff with their plausibility for readers.
\end{abstract}

\section{Introduction}\label{sec:intro}

Over the last few years, Large Language Models (LLMs) like BERT~\cite{DBLP:conf/naacl/DevlinCLT19} or GPT~\cite{DBLP:conf/nips/BrownMRSKDNSSAA20} have been shown to exhibit a worrying tradeoff between their excellent performances~\cite{DBLP:journals/air/AcheampongNC21} and their lack of explainability~\cite{DBLP:journals/corr/abs-2209-11326}. More specifically, while methodologies and metrics to evaluate the performances of such models are well established, the evaluation of their explainability is still a topic of debate. Various criteria have been put forward in the literature, but their evaluation can be challenging and a formal analysis of their interplay is missing. Among qualitative criteria, the faithfullness and plausibilty of the explanations usually come first. Faithfulness requires that ``\emph{an explanation should accurately reflect the (algorithmic) reasoning process behind the model’s prediction}~\cite{DBLP:conf/kdd/Ribeiro0G16,DBLP:conf/acl/JacoviG20}. Plausibility requires that ``\emph{an explanation should be understandable and convincing to the target audience}~\cite{DBLP:journals/corr/abs-1711-07414,DBLP:conf/acl/JacoviG20}. Among qualitative ones, the sensitivity
of the explanations to different types of variations have been investigated~\cite{DBLP:conf/icml/SundararajanTY17,DBLP:conf/nips/AdebayoGMGHK18}. For example, a natural requirement is that ``\emph{explanations should be sensitive (resp., insensitive) to changes in the input that influence (resp., do not influence) the prediction}''.

\medskip

\begin{figure*}[t]
\begin{center}
\includegraphics[angle=0,width=0.85\textwidth]{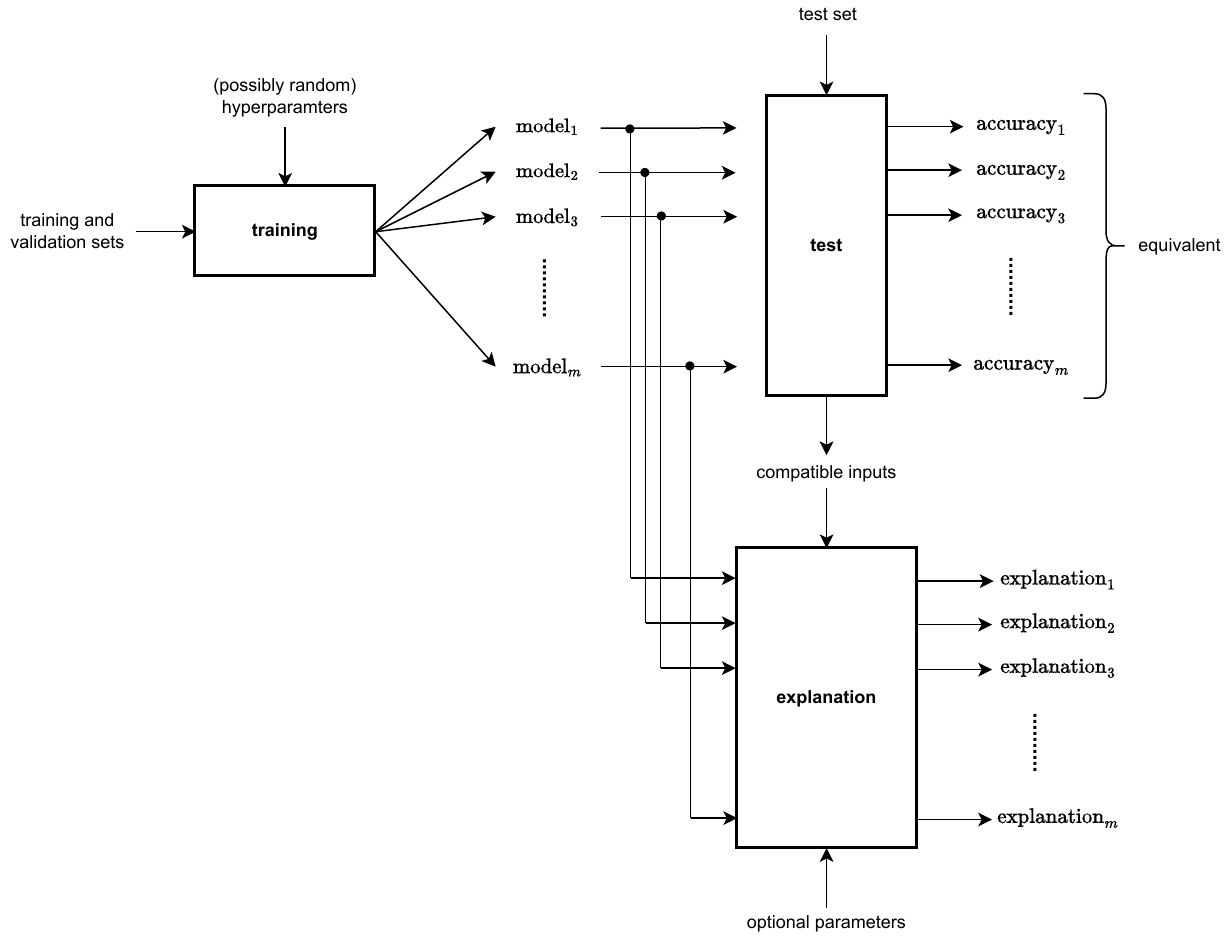}
\caption{Generation of equivalent models and compatible inputs.}\label{fig:blockdiagram}
\end{center}
\end{figure*}

In this paper, we are concerned with the sensitivity of LLMs' explanations to the randomness that can be used in their training, recently put forward in~\cite{TAL_submission} and illustrated in Figure~\ref{fig:blockdiagram}. The starting observation of this previous work is that the explanability of LLMs is usually investigated for fixed models, resulting from a single training. However, the optimization algorithms used to train LLMs may require some randomness (e.g., to initialize the weights of the neurons in a neural network) which may affect their explainability. This randomness is usually not viewed as a useful hyperparameter since the only way to exploit it is to search exhaustively among indistinguishable random seeds. The authors nevertheless observe that in practice, it is sometimes possible to generate many models trained with the same data and different randomness, which lead to similar accuracies.\footnote{~Non-random hyperparameters relying on heuristic selection and leading to equivalent models could be considered as well.} They denote such models as equivalent if their accuracies do not exhibit statistically significant differences. They then identify so-called compatible inputs for which all the equivalent models give the same prediction. This enables the selection of inputs for which no criteria can help deciding whether one model is preferable to another one. 
Finally observing that the explanations of such equivalent models on compatible inputs can significantly differ, they claim that explanations limited to a single model are then insufficient since arbitrary. 
That is, if equivalent models have different explanations on compatible inputs, it is necessary to characterize the explanations' distribution to ensure it is sufficiently informative (e.g., differs from a uniform distribution where all explanations are equally likely). Concretely, the authors propose a visual characterization of the explanations' sensitivity to randomness using box-plots, which is illustrated with a case study of journalistic text classification in French.
For this purpose, they apply a Layerwise Relevance Propagation (LRP) method~\cite{DBLP:conf/cvpr/CheferGW21} to classification results obtained with the CamemBERT model~\cite{DBLP:conf/acl/MartinMSDRCSS20}. They then compare the explanations produced with those of a simpler model exploiting logistic regression and linguistic features~\cite{escouflaire2022}. In both cases, explanations are shaped as attention maps, which are assumed to be easy to understand by human readers~\cite{DBLP:conf/acl/SenHYKR20}.

\medskip

As a natural next step, this paper aims to try pushing these initial investigations towards a more quantitative view. It also posits an assumption in order to help clarifying their impact on the explanations' plausibility. For this purpose, we first need to introduce a taxonomy of explanations. 

\smallskip

First, explanations for the classification of $n$-word texts can display a weight for every word independently or for $t$-tuples of (ordered or un-ordered) words (i.e. display the importance of combination of words). Then, they can be univariate (i.e., provide one attention value per word or tuple of words) or $d$-variate (i.e., provide $d$ such attention values per word or tuple of words). Finally, in case these explanations show a sensitivity to the training randomness of the LLMs, one can characterize this sensitivity with first-order methods (i.e., consider only the means of the explanations' distribution as informative) or higher-order methods (i.e., use higher-order moments of the explanations' distribution).

\smallskip

Based on this taxonomy, it appears that the explanations used in~\cite{TAL_submission} and their analysis are the simplest possible ones: word-level, univariate and first-order. As a result, and inspired by information theory~\cite{DBLP:books/daglib/0016881}, we suggest a quantification of such explanations' sensitivity to the training randomness based on a Signal-to-Noise Ratio (SNR). For this purpose, we (preliminarily) assume that only the mean of word-level univariate  explanations can be understood by a target audience, which we next denote as the \emph{word-level univariate first-order plausibility assumption}, or $(1,1,1)$ plausibility assumption. 
Under this assumption, it is natural to define the signal of the explanations as the variance of the words' attention means and their noise (i.e., the variations of the explanations due to the training randomness) as the mean of the words' attention variances. Such definitions directly lead to the
question whether LLMs can provide ``informative'' explanations measured with the resulting SNR. 
We show in the paper that explanations 
produced with the LRP method for predictions made with the CamemBERT model are actually
less informative than those of the simpler model based on logistic regression and linguistic features.
Hence, our results clarify that unless other tools can provide similarly simple
but more informative explanations, combining the accuracy of LLMs with better explainability
than simpler models based on linguistic features  
may require contradicting the $(1,1,1)$ plausibility assumption and leverage 
more complex explanations, which we discuss in the conclusions of the paper.

\section{Background}\label{sec:background}

\begin{figure*}[t]
    \centering
    \includegraphics[scale = 0.35]{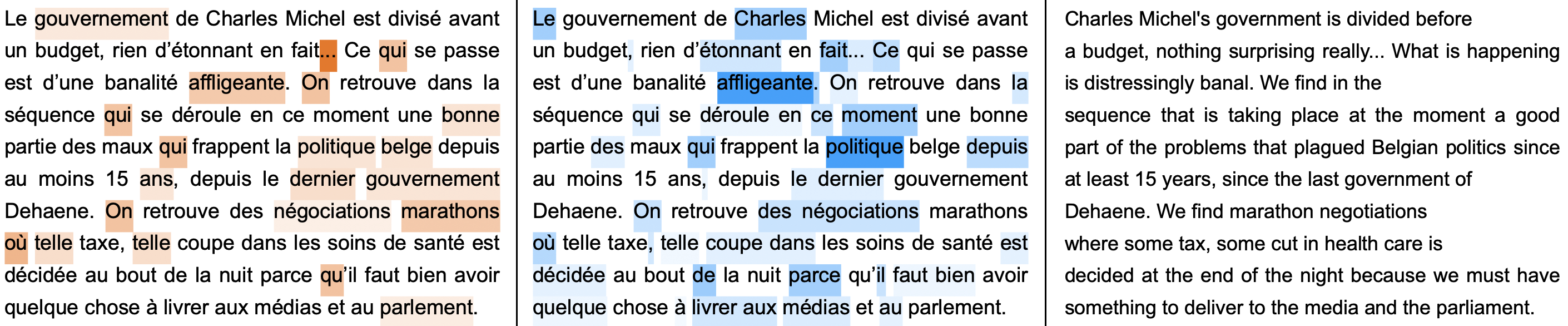}
    \caption{Example of linguistic attention map (left) and LRP attention map (middle) with translation (right).}
    \label{fig:attentionmaps}
\end{figure*}

\subsection{Case study}\label{subsec:case}
We focus on a task of French text classification and use the InfOpinion dataset presented in~\cite{bogaert2023}. The goal of our models is to determine whether a piece of news is categorized as \texttt{information} or \texttt{opinion}. The dataset is balanced and is composed of 5,000 news for the \texttt{information} class (editorials, chronicles, ...) and 5,000 news of the \texttt{information} class (reporting, press dispatch, ...), tackling similar topics. The dataset is split in train, validation and test sets with an 80\%/10\%/10\% ratio. Each set is composed of the same amount of text from each class. 

\subsection{CamemBERT model}\label{subsec:camembert} 
The transformer model we used is called CamemBERT~\cite{DBLP:conf/acl/MartinMSDRCSS20}. It is a French language model based on the same architecture as roBERTa~\cite{DBLP:journals/corr/abs-1907-11692}. It was pre-trained on the French part of the OSCAR dataset (138\,GB of text) by its creators, and we fine-tuned it by training its classification head during our experiment.\footnote{~The pre-trained French model we used is available at the following  address: \url{https://huggingface.co/camembert-base}.} The fine-tuning was done during two epochs. We used a learning rate of $2\times10^{-5}$ and a batch size of 4. The dropout parameter, that allows one to prevent overfitting by de-activating some neurons during the training was set to 10\%. The model takes a random seed as a parameter, usually made public for reproducibility purposes. This seed impacts the order of our training set, the deactivated neurons based on the dropout parameter and the classification head's initial weights. As mentioned in the introduction, we consider it as a random hyperparameter in order to identify equivalent models. 

\subsection{Feature-based model}\label{subsec:feature}
The feature-based model we used is a logistic regression classifier using 18 linguistic features identified as efficient predictors of opinion for French news articles~\cite{escouflaire2022}. Most of these features rely on the ratio of specific tokens in the article: adjectives, verbs, first person pronouns and determiners, relative pronouns, the indefinite third singular pronoun \textit{on}, expressive punctuation signs (semicolons, exclamation marks and question marks), quotation marks, digits, negation words, words longer than 7 characters, words that are in the Lexique3~\cite{new2004} or NRC~\cite{DBLP:journals/ci/MohammadT13} sentiment lexicons. Only two features are not token-related, but global to the whole article: Carroll's corrected type-token ratio~\cite{carroll1964} and the mean word length. We use this  model as an example of classifier that converges towards a unique solution, hence leading to a single explanation for a given text.

\subsection{Explanations}\label{subsec:explanations}

For the transformer model CamemBERT, we generate explanations using
a layerwise relevance propagation approach,
which allows getting word-level attention maps for deep learning models~\cite{bach2015}. It works by back-propagating the relevance from the last layer of the network using conservation constraints, so that the relevance of each neuron is redistributed to the neurons of the previous layer based on their respective gradient. This principle is then followed through the whole network up to the input layer in order to get word-level explanations. As the constraints are more difficult to satisfy for some layers in the models~\cite{DBLP:conf/icml/AliSEMMW22}, this method can be improved 
with additional rules. In this paper, we used an improved version 
from~\cite{DBLP:conf/cvpr/CheferGW21} and refer to it as LRP. 

\medskip

For the feature-based model, we generate word-level explanations using the linguistic attention maps proposed in~\cite{TAL_submission}.
In this case, words are simply highlighted in a brighter tone if they belong to multiple features and if these features have a high coefficient in the regression. Since this approach is restricted to the explanation of word-level features, the method cannot reflect the impact of features related to the entire text. Yet, since the feature-based model only contains two of them (the type token ratio and the average word length), we consider that this explanation method is sufficiently faithful for our evaluations.

\medskip

\noindent Examples of attention maps are given in Figure~\ref{fig:attentionmaps}. These attention maps display word-level univariate explanations.

\section{Experimental setting}\label{sec:setting}

We started our experiment by fine-tuning the CamemBERT model on our training set (8,000 news) with the parameters described in the background section. More precisely, in order to study the impact that changes in the training
randomness have on the explanations of equivalent models, we fine-tuned the same pre-trained model $m$ times by only modifying the random seed used during training. We therefore obtained $m$ versions of our fine-tuned transformer model. For each version, the accuracy was evaluated on a test set of $1,000$ news ($n=1,000$). As in~\cite{TAL_submission}, we then isolated a subset of $m'<m$ most accurate models, so that the difference between the best (a) and worst (b) accuracies of the models in the subset was not statistically significant. To evaluate the equivalence of the models in a subset, we computed the Z statistic~\cite{LehmannZtest}, which can be used to detect whether
two proportions (here, the accuracies $a$ and $b$) are significantly different:
\begin{equation}
    z = \displaystyle\left\lvert\frac{a-b}{\sqrt{\frac{\frac{a+b}{2}*(1-\frac{a+b}{2})}{n}}}\displaystyle\right\rvert
\end{equation}
We considered that $z$ values greater than $1.96$ ($p<0.025$) mean that the accuracies of the best and the worst models in a subset are different. 
For lower $z$ values, we conclude that these accuracies do not differ significantly and therefore, we consider the models in the subset
as equivalent from the performance viewpoint. Concretely, and starting from $m=200$ models, a restriction to $m'=100$ was enough for this.\footnote{~The complexity of finding equivalent models depends on the size of test set and the target $p$-value when evaluating 
their differences in accuracy: the larger the test set, the more confident the estimation of the accuracy so that even small differences can be significant. But this only implies that the ratio between $m$ and $m'$ increases when the size of the test set and the $p$-value increase.}

In parallel, we trained our feature-based model using a grid search. 
The training was shown to converge towards a single model for which we evaluated the accuracy on the same test set as the CamemBERT model.
As expected, the accuracy of this simpler model ($\approx 89\%$) was slightly below the one of the transformer-based one  ($\approx 96\%$).

\smallskip

The next step is to gather explanations for all our models. For the camemBERT models, we used Chefer's LRP method~\cite{DBLP:conf/cvpr/CheferGW21}. For the feature-based model, we used the aforementioned linguistic attention maps. We explained 20 different texts from our test set: 10 short ones (with less than 50 words) and 10 long ones (with more than 400 words). Each of these texts are compatible (i.e., they are assigned the same class by all the models). As a result, for each of these 20 texts, we obtained 100 attention maps for the 100 versions of the CamemBERT model and one attention map for the feature-based model.

\section{Experimental results}\label{sec:results}

Based on the previous experimental setting, and for each $n$-word text,
we obtain a $100\times n$ matrix of attention maps $\bm{A}_t$ for the $m'=100$ equivalent CamemBERT models, and a $n$-element attention map $A_f$  for the feature-based model.

\smallskip

\begin{figure}[h]
    \centering
    \includegraphics[scale = 0.35]{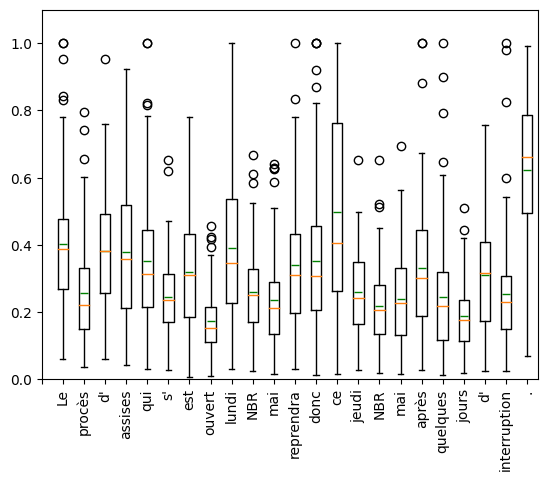}
    \caption{Explanations' box-plot for a transformer model. The green and orange dashes respectively show the mean and the median of the attention distribution for each word. Words with a mean attention value far (resp., close) from 0 are attributed more (resp., less) attention in the explanations in general. Tight boxplots (e.g. ``ouvert'') show a low variability in the attention given by equivalent models. On the contrary, wide boxplots (e.g. ``ce'') show a high variability.}
    \label{fig:boxplot_bert}
\end{figure}

\begin{figure}[h]
    \centering
    \includegraphics[scale = 0.35]{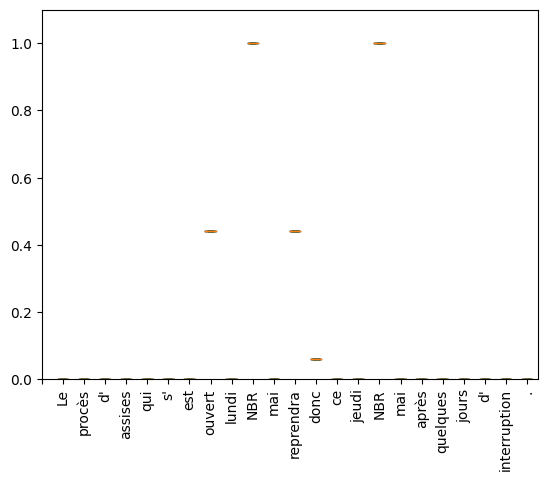}
    \caption{Explanation's box-plot for a feature-based model. As there is no variability in the attention attributed by the feature based model, each boxplot is a simple line. Words that are not used in any feature have an attention score of 0.}
    \label{fig:boxplot_ling}\vspace*{-0.3cm}
\end{figure}

As a first step, we report the box-plot produced from the attention maps of an
exemplary (short) text for our 100 equivalent CamemBERT models in Figure~\ref{fig:boxplot_bert}. For comparison purposes, we also report a similar plot for the single explanations obtained for the feature-based model in Figure~\ref{fig:boxplot_ling}.
One can observe from these plots that by using a number of
equivalent model $m'=100$, the explanations of the CamemBERT give attention to a large amount of words.

\smallskip

Our limitation to word-level, univariate and first-order explanations, formalized as the $(1,1,1)$ plausibility assumption, implies that the only parts of Figure~\ref{fig:boxplot_bert} that is assumed to be understandable by a reader are the green dashes that represent the mean attention values. Those values are given by $\bar{A_t}=\hat{\mathsf{E}}\big(\bm{A}_t\big)$,
with $\hat{\mathsf{E}}$ the sample mean computed over the 100 models so that $\bar{A_t}$ is a
$n$-element vector, like $A_f$.
Under this restriction, it is natural to assume that explanations are (statistically) more informative when the attention of the words vary significantly, leading to a definition of signal:
\begin{equation}
S=\underset{\mathrm{words}}{\hat{\mathsf{Var}}}\left(\underset{\mathrm{models}}{\hat{\mathsf{E}}}\big(\bm{A}_t\big) \right),
\end{equation}
with $\hat{\mathsf{Var}}$ the sample variance computed over the $n$ words of the text, and a definition of noise given by:
\begin{equation}
N=\underset{\mathrm{words}}{\hat{\mathsf{E}}}\left(\underset{\mathrm{models}}{\hat{\mathsf{Var}}}\big(\bm{A}_t\big) \right).
\end{equation}

The SNR is just defined as the ratio between both. In case of explanations that are not sensitive to the training randomness, as with the feature-based model, the noise is null and only the signal can be computed as $\hat{\mathsf{Var}}(A_f)$. 
All these quantities being computed from a limited number of words and models, we additionally evaluate the quality of our estimates. For the signal of the feature-based model we use a standard 95\% confidence interval for the variance. For the signal, noise and SNR of the transformer-based model, we use a 95\% bootstrap confidence interval~\cite{DBLP:books/sp/EfronT93}, which allows dealing with the fact that these quantities are means or variances of estimated vectors.

\medskip

\begin{figure*}[h!]
\centering
\begin{subfigure}{.49\textwidth}
  \centering
  \includegraphics[width=0.9\linewidth]{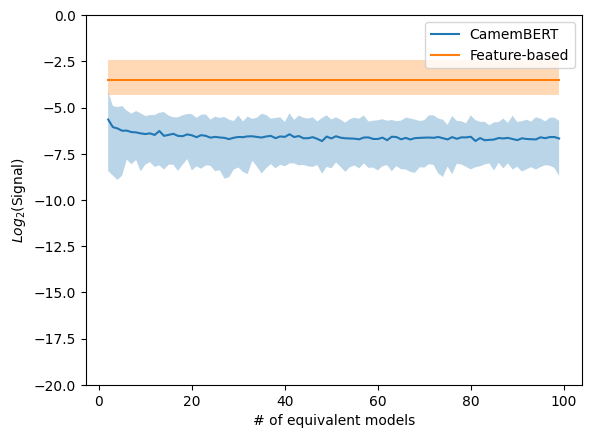}
  \caption{Short text, raw explanations, signal.}\label{fig:SRS}
\end{subfigure}
\begin{subfigure}{.49\textwidth}
  \centering
  \includegraphics[width=0.87\linewidth]{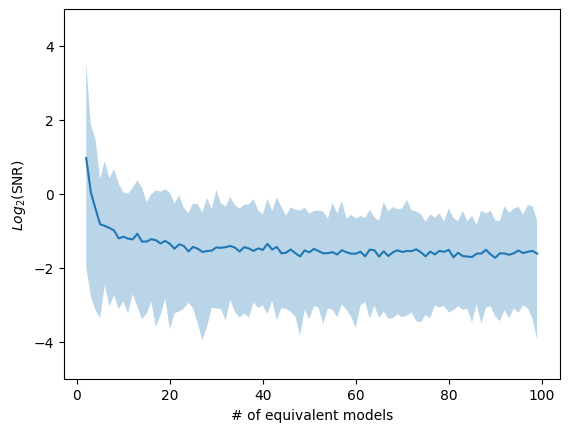}
  \caption{Short text, raw explanations, SNR.}\label{fig:SRSNR}
\end{subfigure}

\begin{subfigure}{.49\textwidth}
  \centering
  \includegraphics[width=0.9\linewidth]{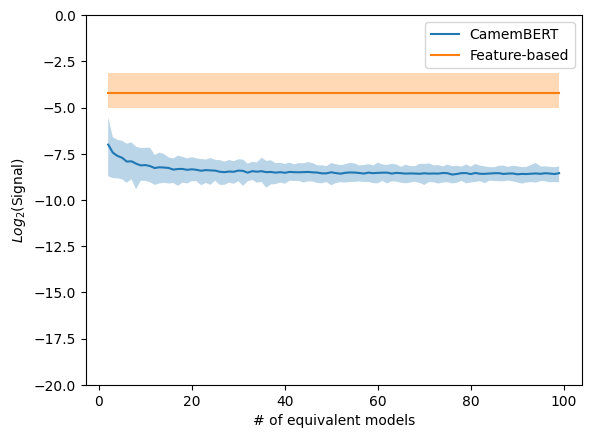}
  \caption{Long text, raw explanations, signal.}\label{fig:LRS}
\end{subfigure}
\begin{subfigure}{.49\textwidth}
  \centering
  \includegraphics[width=0.87\linewidth]{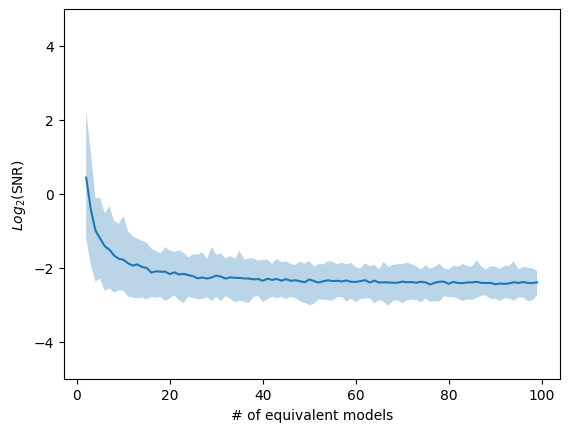}
  \caption{Long text, raw explanations, SNR.}\label{fig:LRSNR}
\end{subfigure}

\begin{subfigure}{.49\textwidth}
  \centering
  \includegraphics[width=0.9\linewidth]{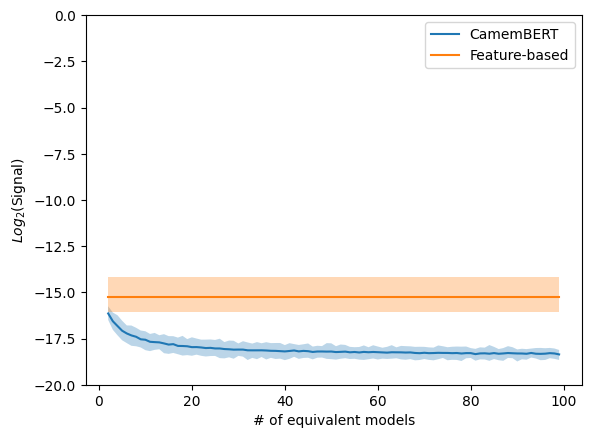}
  \caption{Long text, normalized explanations, signal.}\label{fig:LNS}
\end{subfigure}
\begin{subfigure}{.49\textwidth}
  \centering
  \includegraphics[width=0.87\linewidth]{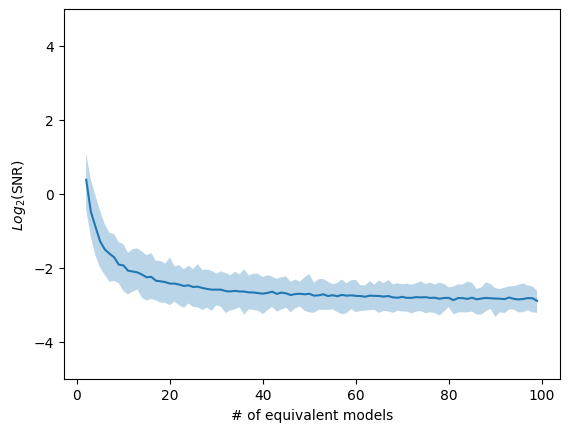}
  \caption{Long text, normalized explanations, SNR.}\label{fig:LNSNR}
\end{subfigure}

\caption{Metrics estimation for illustrative texts.}\label{fig:estimations}
\end{figure*}

We now discuss the main observations that can be extracted from the 
estimation of these quantities given in Figure~\ref{fig:estimations} for exemplary texts,
with the important cautionary note that the definition of signal we use is a statistical one. The possible gap between this definition of signal with other, more semantic, ones will be discussed in conclusions.

\smallskip

Starting with Figures~\ref{fig:SRS} and \ref{fig:SRSNR} vs. \ref{fig:LRS}
and \ref{fig:LRSNR} which represent the signal (of both the CamemBERT and feature-based models) and the SNR (of the CamemBERT models) for a short and long text, a preliminary observation is that the confidence intervals naturally get tighter with longer texts and more equivalent models (which allows improving the estimation of the signal, the noise and therefore the SNR).

\smallskip

Following with the signal estimations in Figures~\ref{fig:SRS} and \ref{fig:LRS}, we can observe two main trends. First, the signal of the CamemBERT models decreases when the number of equivalent models used in its estimation increases. This is because as this number of equivalent models increases, the aggregated explanations tend to cover more words (as observed in Figure~\ref{fig:boxplot_bert}) and their average tends to be flattened. Second, the (average) signal of the CamemBERT models is always lower than the one of the feature-based model. This essentially suggests that when limited to a simple (first-order) analysis of simple (word-level, univariate) explanations, as
formalized by the $(1,1,1)$ plausibility assumptions, the simpler feature-based model tends to provide more informative explanations than the CamemBERT ones, in the statistical sense captured by our definitions of signal and noise.

\smallskip

In order to try reducing the impact of the more disparate assignment of weights in the explanations of the CamemBERT models with the LRP method, we additionally tried to post-process them. For example, Figure~\ref{fig:LNS} shows the signal of normalized explanations, where we ensure that the explanations of the CamemBERT and feature-based models have the same number of words with non-zero weights and that these weights sum to one (which reduces the faithfulness of the CamemBERT explanations).
Compared to Figure~\ref{fig:LRS}, this naturally reduces the explanations' signal for all models and makes them closer. Yet, the average signal of the CamemBERT models remain substantially lower.

\smallskip

Eventually, the easiest to interpret observations are obtained from the SNR Figures~\ref{fig:SRSNR} and~\ref{fig:LRSNR}, which are both reaching values below one. This implies that the variance of the weights assigned by the individual explanations of the CamemBERT models to each word, computed over the training randomness of the equivalent models, is actually larger than the variance of the average weights computed over the words of a text, which we assume represents the informativeness of the explanations. The concrete value reached ($\approx 0.25$) also gives an indication of how many equivalent models must be produced in order to reach a good estimation of the signal 
(i.e., one must typically average the explanations so that the noise of the average explanations becomes smaller than the one of raw explanations).
This averaging requirement is increasingly undesirable for models of which the optimization is computationally-intensive.

\section{Conclusion}\label{sec:conclusion}

Our results highlight that if limited to simple explanations, simple models may carry more statistical signal (i.e., produce explanations putting attention on fewer words with more contrast) and less noise (i.e., be less sensitive to the training randomness) than transformer-based models
like CamemBERT. Unless other simple explanation tools lead to different conclusions, it means that combining such LLMs with as good explainability as simple models may require more complex explanations. 
Admittedly, this conclusion is based on a single 
classification task. Assessing it with other datasets
is therefore needed to evaluate its extent. More generally, and
in all the contexts where it holds, this conclusion is based on a statistical definition of signal and noise which may deviate from a more semantic definition. As a result, our results are also calling to challenge the 
$(1,1,1)$ plausibility assumption and to investigate whether human readers can understand more complex $(t,d,o)$ explanations
considering $t$-tuples of words, $d$ attention values per word and an higher-order statistics, which could all contribute to improve the explainability of LLMs. Explanations based on $t$-tuples of (ordered or un-ordered) words could improve the faithfulness of the explanations and make them more informative since it is likely that the improved accuracy of LLMs is taking advantage of such combinations of words. Explanations based on $d$ attention values per word could have a similar impact, since they would avoid aggregating the weights that different explanation features add to each word. Finally, moving to higher-order evaluations could lead to transform a part of the variance that we currently capture as noise into useful signal. This would for example take place if the variance of the explanations observed was corresponding to a variance of viewpoints, similar to the one that could be observed for human annotators. Investigating this question would be an interesting follow up work, given that our subjectivity detection task would likely lead to different readers having different opinions on which words are important to explain 
a classification. Ultimately, our results therefore re-emphasize the fundamental question whether accuracy and explainability are unavoidably the result of a tradeoff, or if one can combine the excellent accuracy of recent LLMs with equally plausible explanations as simpler feature-based models.

\medskip
\noindent \textbf{Acknowledgments.} François-Xavier Standaert is Senior Research Associate of the F.R.S.-FNRS. Work supported by the Service Public de Wallonie Recherche, grant number 2010235-ARIAC by DIGITALWALLONIA4.AI.

\bibliography{refs}
\end{document}